\documentclass[9pt, conference]{IEEEtran}
\IEEEoverridecommandlockouts

\newcommand{\R}{\mathbb{R}}

\newcommand{\W}{\boldsymbol{W}}

\newcommand{\x}{\boldsymbol{x}}
\newcommand{\y}{\boldsymbol{y}}

\renewcommand{\H}{\boldsymbol{H}}
\newcommand{\J}{\boldsymbol{J}}
\renewcommand{\b}{\boldsymbol{b}}

\usepackage{cite}
\usepackage{amsmath,amssymb,amsfonts}
\usepackage{graphicx}
\usepackage{textcomp}
\usepackage{xcolor}
\usepackage{fancyhdr}
\usepackage{amssymb, booktabs, mathtools, amsthm}
\usepackage{hyperref}
\usepackage{tcolorbox}
\usepackage{subcaption}
\usepackage{algorithm}
\usepackage{algpseudocode}
\usepackage{colortbl}
\usepackage{xcolor}
\usepackage{multirow}
\definecolor{green}{RGB}{200, 230, 201} 

\def\x{{\mathbf x}}
\def\b{{\mathbf b}}
\def\W{{\mathbf W}}

\def\J{{\mathbf J}}
\def\H{{\mathbf H}}

\def\y{{\mathbf{y}}}
\def\R{\mathbb{R}}

\usepackage{color}

\mathtoolsset{showonlyrefs}

\def\BibTeX{{\rm B\kern-.05em{\sc i\kern-.025em b}\kern-.08em
    T\kern-.1667em\lower.7ex\hbox{E}\kern-.125emX}}
\makeatletter
\newcommand{\linebreakand}{%
  \end{@IEEEauthorhalign}
  \hfill\mbox{}\par
  \mbox{}\hfill\begin{@IEEEauthorhalign}
}
\makeatother

\begin{document}

\title{\huge {GradNetOT}: Learning Optimal Transport Maps with GradNets}

\author{\IEEEauthorblockN{Shreyas Chaudhari\IEEEauthorrefmark{1} \qquad
Srinivasa Pranav\IEEEauthorrefmark{1} \qquad
Jos\'e M. F. Moura}
\IEEEauthorblockA{Electrical and Computer Engineering, Carnegie Mellon University\\
\{\texttt{shreyasc@andrew.cmu.edu}, \texttt{spranav@cmu.edu}, \texttt{moura@andrew.cmu.edu} \}}
\thanks{\IEEEauthorrefmark{1}Authors contributed equally. Authors partially supported by NSF Graduate Research Fellowships (DGE-1745016, DGE-2140739), NSF Grant CCF-2327905, and an ARCS Fellowship. Authors thank Kawisorn Kamtue for helpful discussions. Code available at \href{https://github.com/cShreyas/GradNetOT}{https://github.com/cShreyas/GradNetOT}. 

{\textcopyright~2025 IEEE. Personal use of this material is permitted. 
Permission from IEEE must be obtained for all other uses, in any current or future media, 
including reprinting/republishing this material for advertising or promotional purposes, 
creating new collective works, for resale or redistribution to servers or lists, 
or reuse of any copyrighted component of this work in other works.}}
}
\maketitle
\begin{abstract}
Monotone gradient functions play a central role in solving the Monge formulation of the optimal transport (OT) problem, which arises in modern applications ranging from fluid dynamics to robot swarm control. When the transport cost is the squared Euclidean distance, Brenier’s theorem guarantees that the unique optimal transport map satisfies a Monge-Amp\`ere equation and is the gradient of a convex function. In \cite{MGNsICASSP, GradNetsTSP}, we proposed Monotone Gradient Networks (mGradNets), neural networks that directly parameterize the space of monotone gradient maps. In this work, we leverage mGradNets to directly learn the optimal transport mapping by minimizing a training loss function defined using the Monge-Amp\`ere equation. We empirically show that the structural bias of mGradNets facilitates the learning of optimal transport maps across both image morphing tasks and high-dimensional OT problems.
\end{abstract}


\section{Introduction}
The optimal transport (OT) problem originated with the work of Gaspard Monge in 1781, who was interested in a fundamental problem that arises when constructing fortifications: how can one transfer molecules of soil from piles to holes while minimizing the cost of transfer~\cite{monge1784m}? Since then, Monge's optimal transport problem and Leonid Kantorovich's relaxation~\cite{kantorovich1942translocation,kantorovich1948problem} have been phrased in terms of probability densities and finding an optimal transport map: how can a source probability distribution be mapped to a target probability distribution in a way that minimizes the transport cost?

These formulations of optimal transport, along with several newer variations, have found applications spanning not only statistics and logistics, but also signal and image processing, robotics, machine learning, fluid dynamics, economics, geology, astronomy, and beyond. In our simulations, we illustrate an example of optimal transport for image morphing applications.
Robotics applications of OT include the control of large swarms of robots or unmanned aerial vehicles (UAVs)~\cite{alqudsi2025uav}, which have applications ranging from critical natural disaster rescue missions to intricate and artistic light shows. By modeling movements of large robot swarms as time-evolving spatial probability distributions,  an optimal transport map can guide the swarm from an initial distribution to a desired target distribution while minimizing travel cost~\cite{SwarmAsDist_ma2023high}.
Furthermore, OT applications to economics include works showing the equivalence between optimal transport, stable matching, and modeling hedonic price equilibria, which capture relationships between prices and product characteristics~\cite{chiappori2010hedonic}. In geology, developments concerning OT on Riemannian manifolds have been used to model continental drift by describing ground mass distributions and geodesic transport paths over time~\cite{cohen2021riemannian}. In astrophysics, recent works on Monge-Amp\`ere gravity employ optimal transport maps to model the distribution of mass throughout the universe and the formation of cosmological structures~\cite{MAG_bonnefous2024monge}. 

On the theory side, in 1991, Yann Brenier proved a key result that became a catalyst for optimal transport flourishing into an active research field in the modern era~\cite{brenier1991polar}. Brenier showed the equivalence between the Monge optimal transportation problem and the elliptic Monge–Amp\`ere equation, leading to the following result: when the source and target probability distributions are well-defined and the transport cost is the squared Euclidean distance, then the optimal transport map is the gradient of a convex potential function that solves the corresponding Monge-Amp\`ere equation.

Exploiting Brenier's link between optimal transport and the Monge-Amp\`ere equation is nontrivial because the Monge-Amp\`ere equation is highly non-linear and generally difficult to solve using standard techniques like calculus of variations~\cite{thorpe2018introductionOT}. While some works, such as \cite{MAG_bonnefous2024monge}, rely on additional assumptions like spherical symmetry to recover an analytic solution, the majority of the literature uses numerical approaches like discrete-discrete entropic regularization and semi-discrete optimal transport~\cite{peyre2019computational}. In this paper, we focus on the intersection of Brenier's result and advancements in neural networks to solve Monge optimal transport problems.

We propose GradNetOT: a method that leverages monotone gradient networks (mGradNets)~\cite{GradNetsTSP} to solve Monge optimal transport problems. Using an mGradNet to parameterize the optimal transport map, we search through the space of all gradients of convex potential functions. To guide our search, we explicitly use the Monge-Amp\`ere equation in our optimization objective.

\section{Background}
We briefly review probability concepts relevant to optimal transport (OT). A \emph{probability measure} $\mu$ assigns mass to events, which are measurable subsets of a sample space $\mathcal{X}$. The probability measure satisfies the Kolmogorov axioms of non-negativity, countable additivity, and normalization ($\mu(\mathcal{X}) = 1$). The probability of an event $A \subseteq \mathcal{X}$ is expressed as $\mu(A) = \int_A d\mu(\x)$. For example, if $\mu$ is discrete, it assigns probability mass to a finite set of points and $\mu(A) = \sum_{x \in A} \mu(\x)$. Alternatively, if $\mathcal{X} = \R^N$, then the Lebesgue measure yields the $n$-dimensional volume of $A\subset \R^N$. If $\mu$ is absolutely continuous with respect to the Lebesgue measure, then it admits a probability density function $p(\x)$ and $\mu(A) = \int_{A} p(\x) dx$.

Probability measures can be transformed by functions. Given a function $f:\mathcal{X} \to \mathcal{Y}$, where $\mathcal{X}$ is equipped with probability measure $\mu$, the \emph{pushforward measure} $f_\# \mu$ is the resulting probability measure on $\mathcal{Y}$ induced by $f$. For $A \subseteq \mathcal{Y}$, the pushforward is defined by:
\begin{align}
    f_{\#}\mu(A) = \mu(f^{-1}(A)) =  \mu(\{\x \in \mathcal{X} : f(\x) \in A\}) \label{eq:pushforward_intro}
\end{align}
where $f^{-1}(A)$ corresponds to the preimage of $A$ under $f$. Intuitively, \eqref{eq:pushforward_intro} computes the probability of set $A$ in the target space $\mathcal{Y}$ by measuring the probability of the set in the source space $\mathcal{X}$ whose image under $f$ is $A$. Additionally, $f$ is called a diffeomorphism if $f$ is a bijection and both $f$ and $f^{-1}$ are differentiable.

\section{Problem Statement}
Optimal transport involves finding the most efficient way to transform one probability distribution into another. The distributions can be thought of as descriptions of how mass (representing probability) is spread over a space. Efficiency here is measured by a predefined cost function.

Formally, let $\mathcal{X}$ and $\mathcal Y$ be spaces equipped with respective probability measures $\mu$ and $\nu$. Let $c: \mathcal{X} \times \mathcal{Y} \to [0, \infty)$ be a cost function, where $c(\x, \y)$ denotes the cost of moving a unit of mass (probability) from $x \in \mathcal{X}$ to $y \in \mathcal{Y}$. The optimal transport problem is then to find a map $T: \mathcal{X} \to \mathcal{Y}$ that pushes forward $\mu$ to $\nu$ (i.e., $T_\#\mu = \nu$), and minimizes the total transport cost:
\begin{align}
\inf_{T: T_{\#}\mu = \nu} &\int_{\mathcal{X}} c(\x, T(\x)) d\mu(\x) \label{eq:general_ot}
\end{align}
The formulation in~\eqref{eq:general_ot} is known as the Monge problem and applies to settings where the probability measures $\mu$ and $\nu$ are either discrete or continuous~\cite{thorpe2018introductionOT}. In general, establishing the uniqueness, and structure of the optimal transport map $T$ that solves~\eqref{eq:general_ot} is challenging~\cite{de2014monge}. In fact, a map satisfying $T_{\#}\mu = \nu$ may not exist at all since the Monge problem does not permit mass splitting.  For example, consider discrete measures where $\mu$ assigns all its mass to a single point, while $\nu$ distributes the mass evenly between two distinct points. Since $T$ is a deterministic map sending each point $\x \in \mathcal{X}$ to a single point $\y\in\mathcal{Y}$, no such $T$ can push $\mu$ forward to $\nu$~\cite{thorpe2018introductionOT}.

The existence and uniqueness of the solution to the Monge problem in~\eqref{eq:general_ot} can be established with additional assumptions on the probability measures and cost function~\cite{de2014monge}. In this work, we specify $\mathcal{X} = \mathcal{Y} = \R^N$ and assume $T$ is a diffeomorphism.\footnote{See Theorem 3.6 in~\cite{de2014monge} for generalizations of this setting.}
Brenier's theorem \cite{brenier1991polar, santambrogio2015optimal} guarantees that if the probability measures $\mu$ and $\nu$ are absolutely continuous with respect to the Lebesgue measure, i.e., they admit respective probability densities $p(\x)$ and $q(\y)$, and the cost function is squared Euclidean distance, the solution to~\eqref{eq:cont_ot} is unique and can be expressed as 
\begin{equation}
    T = \nabla \phi \label{eq:grad_of_convex}
\end{equation}
where $\phi : \R^N \to \R$ is a convex potential function and $\nabla \phi: \R^N \to \R^N$ denotes its monotone gradient. Furthermore, as proved in \cite{rockafellar1970convex}, $T$ being the monotone gradient of $\phi$ means $T$ is cyclically monotone. This corresponds to the intuition that the transport routes taken by units of mass (probability) do not cross or intersect, as detailed in Sec. IX-A of \cite{MAG_bonnefous2024monge}. Moreover, in this setting, $T$ is smooth almost everywhere~\cite{de2014monge}. As described in~\cite{benamou2000computational, de2014monge}, $T$ being smooth and one-to-one means that the constraint $T_{\#}\mu = \nu$ can be expressed in terms of the corresponding densities as
\begin{align}
    \int_{T^{-1}(A)} p(\x) d\x &= \int_{A} q(\y) d\y = \int_{T^{-1}(A)} q(T(\x)) |\det \J_T(\x)| d\x
\end{align}
where $\J_T(x)$ denotes the $N \times N$ Jacobian of $T$ evaluated at $\x$ and where the second equality follows from the change of variables formula applied to the substitution $\y=T(\x)$. As described in \cite{brenier1991polar, MAG_bonnefous2024monge}, this yields the following standard pushforward expression: 
\begin{align}
    p(\x) = q(T(\x))|\det \J_T(\x)|
    \label{eq:pushforward_expr}
\end{align}
Substituting for $T = \nabla \phi$ reveals that the convex potential $\phi$ satisfies a type of Monge-Amp\`ere equation \cite{brenier1991polar,benamou2000computational,MAG_bonnefous2024monge}
\begin{align}
    p(\x) &= q(\nabla\phi(\x)) \det(\H_\phi(\x))
    \label{eq:monge_ampere}
\end{align}
where $\H_\phi = \J_T$ is the $N\times N$ positive semidefinite Hessian of $\phi$.

In summary, for Monge optimal transport problems with squared Euclidean cost (satisfying appropriate regularity conditions), the optimal transport map $T$ exists and is unique. In this work, we specifically leverage \eqref{eq:grad_of_convex} and \eqref{eq:pushforward_expr} to learn $T$.

\section{Related Work}
Brenier's theorem has inspired several approaches for training neural networks to learn the optimal transport mapping between continuous measures with squared Euclidean cost. For example, \cite{ConvexPotentialFlows, makkuva2020optimal} first parameterize a convex potential $\phi$ using an input convex neural network (ICNN) \cite{ICNN} and train the gradient of the ICNN with respect to the input to approximate the optimal transport map. Unfortunately, training the gradient maps of ICNNs often leads to numerical optimization challenges \cite{bunne2022supervised, ICGN, korotin2021neural}.

Other works have considered parameterizing the monotone gradient map without first characterizing the underlying potential. For example, \cite{ICGN} parameterize a Gram factor of $\H_\phi$, the positive semidefinite Hessian of the potential $\phi$, and obtain the gradient map by numerically integrating the learned Hessian. The only known architecture suitable for parameterizing the Gram factor is a single-layer network of the form $\sigma(\W\x + \b)$, which does not require numerical integration and serves primarily as a proof of concept. Moreover, the reliance on numerical integration poses significant challenges for optimal transport in high-dimensional settings.

In contrast, we proposed neural network architectures that are guaranteed to directly model monotone gradients of convex functions in \cite{MGNsICASSP}. Then, in \cite{GradNetsTSP}, we proved that these neural networks are universal approximators of monotone gradient functions and introduced mGradNets that offer expanded parameterizations more amenable for optimization in practice. In this work, we leverage the Monge-Amp\`ere equation to design a principled approach for training mGradNets to solve Monge optimal transport problems where both the source and target probability densities are known. This contrasts with our approach in \cite{MGNsICASSP}, where the source probability density was unknown and we trained MGNs (predecessors of mGradNets) by maximizing the target likelihood of the mapped samples -- the approach taken previously by~\cite{ConvexPotentialFlows}.

\section{Proposed Method}
We propose GradNetOT, a method for computing the optimal transport map in~\eqref{eq:general_ot} with squared Euclidean cost and continuous probability measures $\mu$ and $\nu$. Specifically, we are interested in finding $T$ that minimizes the total transport cost:
\begin{align}
    \inf_{T:T_{\#}\mu = \nu} &\int_{\mathcal{X}} \|\x - T(\x)\|^2 d\mu(\x)
    \label{eq:cont_ot}
\end{align}
As a result of Brenier's theorem, the optimal transport map $T$ solving~\eqref{eq:cont_ot} has the following properties: 1) $T$ has a symmetric positive semidefinite Jacobian since it is the monotone gradient of a convex function; 2) $T$ satisfies the pushforward expression in~\eqref{eq:pushforward_expr} that is closely related to the Monge-Amp\`ere equation. Hence, we propose to parameterize $T$ as a monotone gradient network (mGradNet) \cite{MGNsICASSP, GradNetsTSP}, and train it to satisfy the constraint in~\eqref{eq:pushforward_expr}. This leverages Brenier's theorem~\cite{brenier1991polar}: a function that is the gradient of a convex function and satisfies the constraint must be the unique optimal transport map. 

\begin{figure*}[htpb]
    \centering
    \begin{subfigure}[b]{0.19\textwidth}
        \includegraphics[width=\linewidth, trim=0 0 0 25, clip]{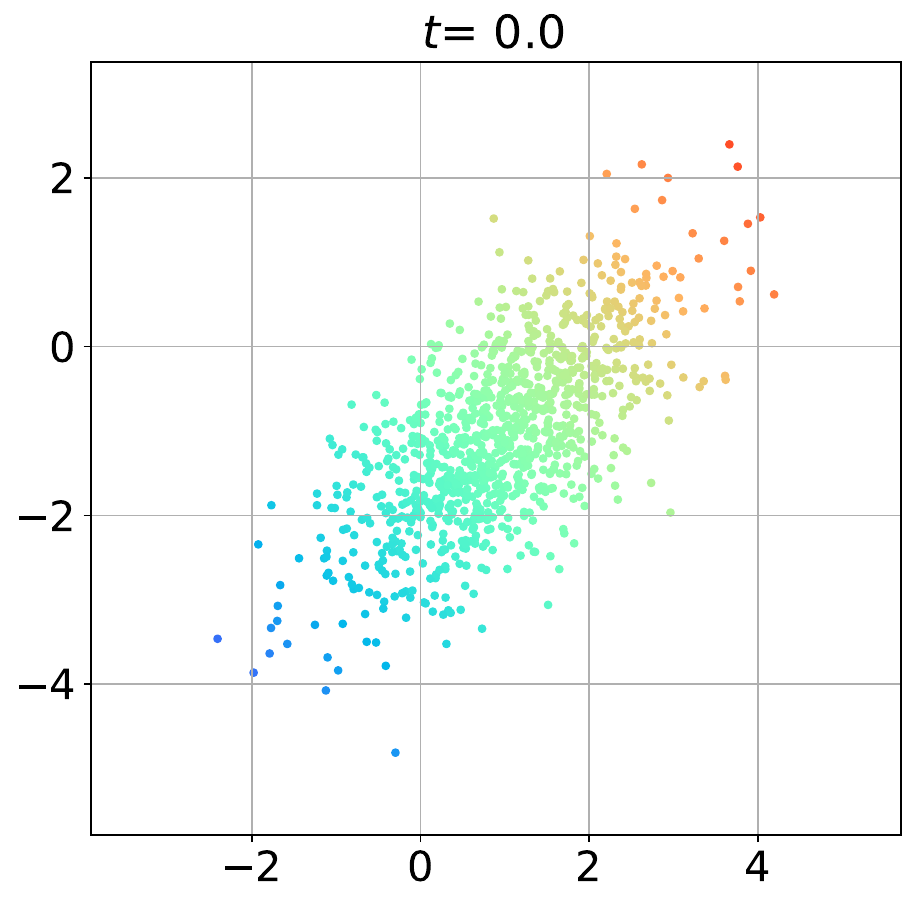}
        \caption{Source Samples}
        \label{subfig:source}
    \end{subfigure}
    \begin{subfigure}[b]{0.19\textwidth}
        \includegraphics[width=\linewidth, trim=0 0 0 25, clip]{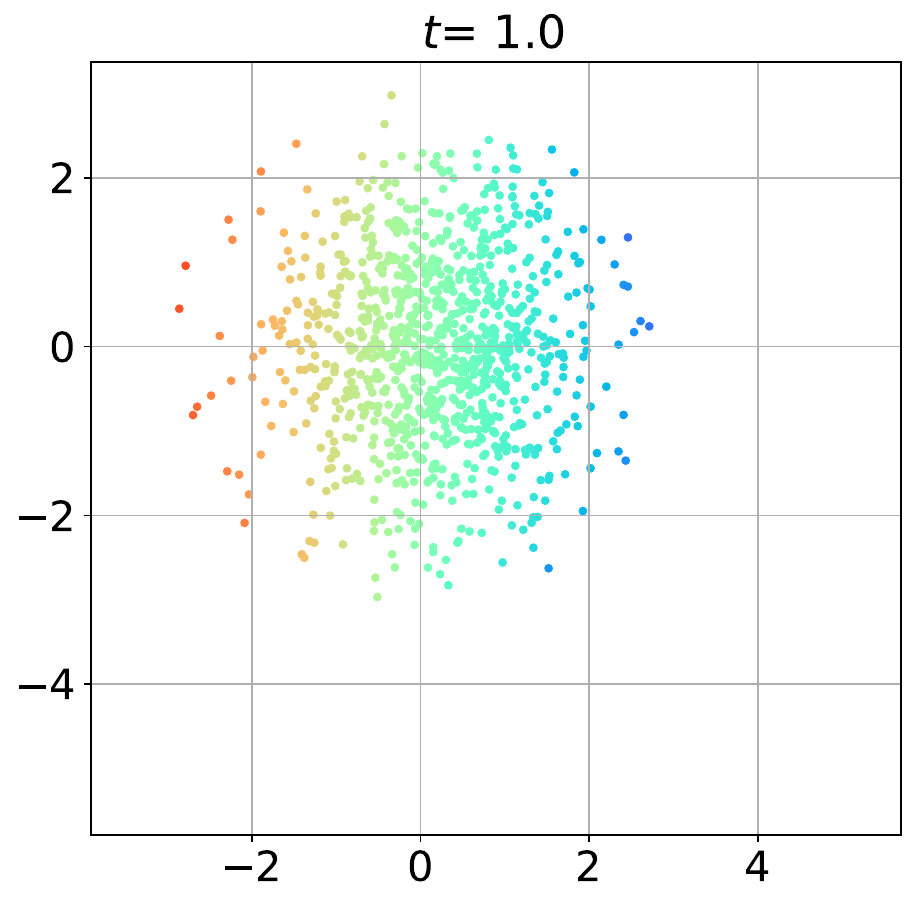}
        \caption{Baseline}
        \label{subfig:baseline}
    \end{subfigure}
    \begin{subfigure}[b]{0.19\textwidth}
        \includegraphics[width=\linewidth, trim=0 0 0 25, clip]{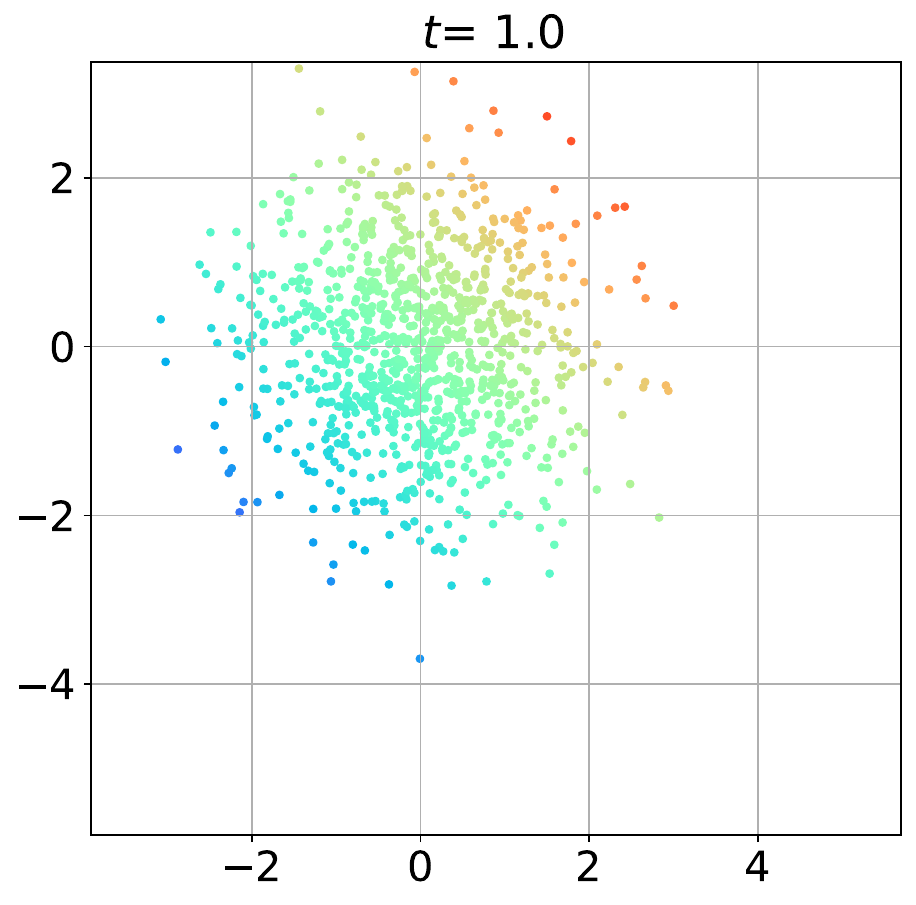}
        \caption{mGradNet-C}
    \end{subfigure}
    \begin{subfigure}[b]{0.19\textwidth}
        \includegraphics[width=\linewidth, trim=0 0 0 25, clip]{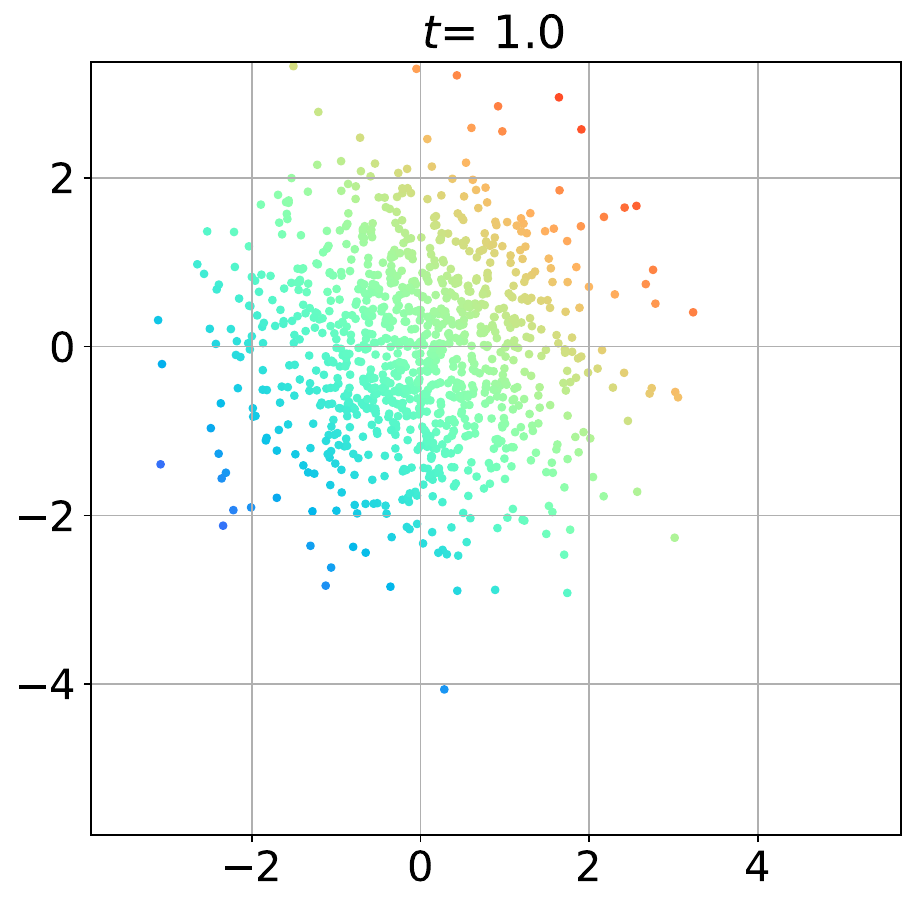}
        \caption{mGradNet-M}
    \end{subfigure}
    \begin{subfigure}[b]{0.19\textwidth}
        \includegraphics[width=\linewidth, trim=0 0 0 25, clip]{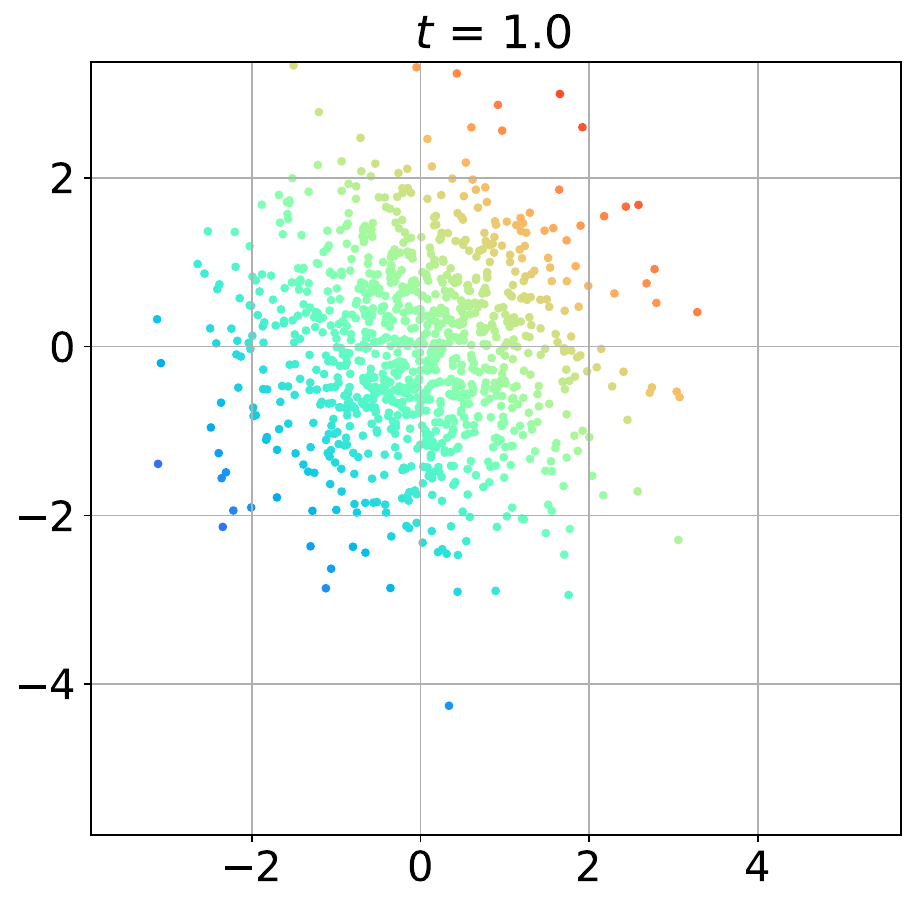}
        \caption{Whitening Transform}
    \end{subfigure}
    \caption{OT from skewed to standard Gaussian. Points are colored based on their positions in the source distribution.}
    \label{fig:gaussians}
\end{figure*}

mGradNets are a class of neural network architectures that are explicitly designed to correspond to monotone gradients of convex functions. The Jacobian of a mGradNet's output with respect to its input is always symmetric positive semidefinite -- for all inputs $\x$ and all parameter values $\boldsymbol{\theta}$ -- thereby guaranteeing that for any mGradNet $T_{\boldsymbol{\theta}}$, there exists a convex potential $\phi_{\boldsymbol{\theta}}$ such that $T_{\boldsymbol{\theta}} = \nabla \phi_{\boldsymbol{\theta}}$. Moreover, mGradNets are \textit{universal approximators} of gradients of convex functions and can be additionally tailored to correspond to subclasses of convex functions, such as convex functions expressible as the sum of convex ridge functions.
We refer to our paper~\cite{GradNetsTSP} for further details and proofs regarding mGradNets and GradNets (a generalization to gradients of $L$-smooth functions). 

\begin{algorithm}
\caption{GradNetOT Training Procedure}
\begin{algorithmic}
\Require mGradNet $T_{\boldsymbol{\theta}}$, source density $p(\x)$, target density $q(\y)$, learning rate $\eta$, batch size $B$, iterations $I$
\vspace{0.2em}

\For{iteration $i = 1$ to $I$}
    \State Sample $\{\x_1,\x_2,\dots,\x_B\} \sim p(\x)$
    \State Compute pushforward points: $\y_j = T_{\boldsymbol{\theta}}(\x_j),\;j=\{1,\dots,B\}$
    \State Compute Jacobians: $\J_{T_{\boldsymbol{\theta}}}(\x_j) = \frac{\partial}{\partial \x_j} T_{\boldsymbol{\theta}}(\x_j)$
    \State Compute loss:
    \[
        \mathcal{L}(\boldsymbol{\theta}) = \frac{1}{B} \sum_{j=1}^B \ell \left( \log \det \J_{T_{\boldsymbol{\theta}}}( \x_j),  \log p(\x_j) - \log q(\y_j) \right)
    \]
    \State Update parameters: \( \boldsymbol{\theta} \gets \boldsymbol{\theta} - \eta \nabla_{\boldsymbol{\theta}} \mathcal{L}(\boldsymbol{\theta}) \)
\EndFor\\
\Return $\boldsymbol{\theta}$
\end{algorithmic}
\label{alg:main_alg}
\end{algorithm}

The training procedure, as provided in Algorithm \ref{alg:main_alg}, is inspired by the training of physics informed neural networks \cite{raissi2019physics} designed to satisfy differential equations. During training, points $\{\x_j\}_{j=1}^B$ are sampled from the source distribution $p$ and transformed by the mGradNet $T_{\boldsymbol{\theta}}$. Using the target density $q$ to evaluate the density of the transformed points leads to computing the training label as $\log p(\x_j) - \log (q(T_{\boldsymbol{\theta}}(\x_j))$. The log determinant of the Jacobian of $T_{\boldsymbol{\theta}}$ (with respect to the input) is evaluated at $\x_j$ and trained to match this label via standard backpropagation. The training procedure presented in Algorithm~\ref{alg:main_alg} minimizes the empirical version of the following objective over the mGradNet parameters $\boldsymbol{\theta}$, where $\ell$ is a loss function like the squared Euclidean distance function:
\begin{align}
    \mathbb{E}_{\x \sim p(\x)} \ell(\J_{T_{\boldsymbol{\theta}}}(\x), \log p(\x) -\log (q(T_{\boldsymbol{\theta}}(\x)))
\end{align}
In contrast to approaches that first characterize the underlying potential only to later extract the gradient, our method directly characterizes the gradient map itself. Parameterizing the underlying convex potential $\phi$, e.g., using an ICNN, and solving~\eqref{eq:monge_ampere} using a generalization of the approach presented would require computing the Hessian of $\phi$ using auto-differentiation, which is much more computationally expensive than computing the Jacobian. By forgoing direct characterization of the potential, our method enables us to consider the expression in~\eqref{eq:pushforward_expr} involving the Jacobian. 

\section{Experiments}
We demonstrate the efficacy of the proposed GradNetOT procedure with a series of optimal transport tasks. In all experiments, we use the Adam optimizer and a fixed batch size of 1000.

\subsection{Gaussians}
We consider the OT problem in~\eqref{eq:cont_ot} with quadratic cost and Gaussian measures $\mu$ and $\nu$. The OT map between multivariate Gaussians $\mathcal{N}(\boldsymbol{\mu}_0, \boldsymbol{\Sigma}_0)$ and $\mathcal{N}(\boldsymbol{\mu}_1, \boldsymbol{\Sigma}_1)$ has an analytic solution. In particular, when the target density is the multivariate standard normal ($\boldsymbol{\mu}_1 = \boldsymbol{0}, \boldsymbol{\Sigma}_1 = \boldsymbol{I}$), the OT map is given by the whitening transform $T(\x) = \boldsymbol{\Sigma}_0^{-1/2}(\x - \boldsymbol{\mu}_0)$. We train the mGradNet-C and mGradNet-M architectures from \cite{GradNetsTSP} using Algorithm~\ref{alg:main_alg}. Using the same method, we also a train a baseline feedforward neural network that is not constrained to be the gradient of a convex function. We visualize the OT maps for the neural network baseline, mGradNet-C, mGradNet-M and the ground truth OT map whitening in Fig.~\ref{fig:gaussians}. The points are colored in the figure according to their corresponding positions in the source distribution. As shown in the figure, the neural network baseline fails to learn the OT map as indicated by the fact that the ordering of the points changes drastically: the red source points are on the right in Fig.~\ref{subfig:source}, but they are mapped to the red points on the left in Fig.~\ref{subfig:baseline}. This indicates that the baseline neural network is not learning a monotone mapping. Meanwhile, the mGradNet-C and mGradNet-M can learn the OT map and their displacements closely match those of the known OT map, namely the whitening transform.

\begin{figure}[h]
    \centering
    \includegraphics[width=0.9\linewidth]{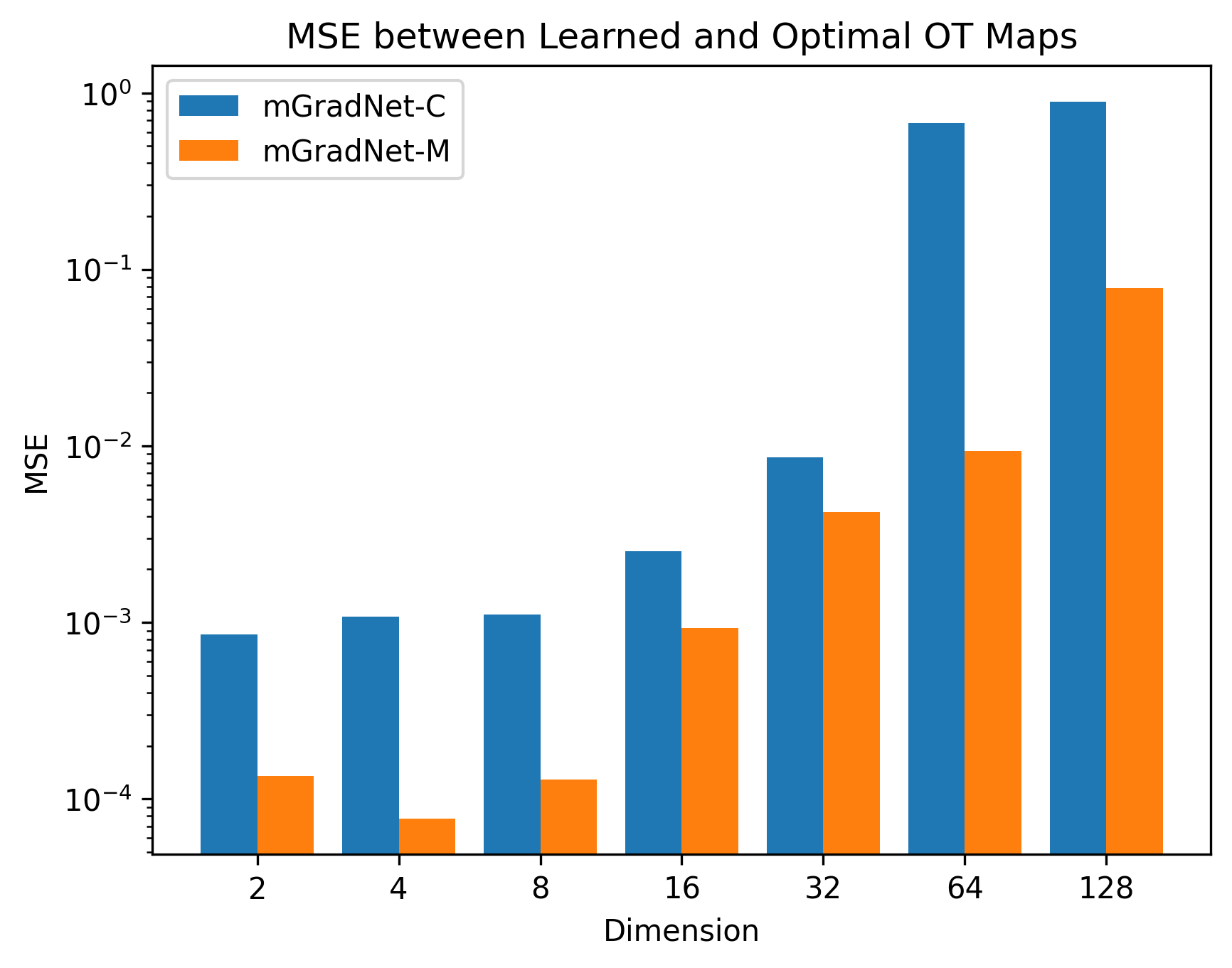}
    \caption{MSE for 1000 samples from skewed multivariate Gaussians mapped to standard Gaussians by mGradNets.}
    \label{fig:high_dim_gauss}
\end{figure}

To test with higher-dimensional Gaussians, we randomly generate a $d$-dimensional source Gaussian distribution by sampling the mean from the standard Gaussian distribution and by sampling the covariance matrix from a Wishart distribution with $d+1$ degrees of freedom and identity scale matrix. After using Algorithm~\ref{alg:main_alg} to train mGradNets to learn the OT map to a standard Gaussian, we evaluate the neural networks on a set of 1000 test points. The whitening transform applied to the test points again yields the ground truth target under the optimal transport map. We use our mGradNets to map the test points and compute the mean squared between them and the ground truth mapped points. Fig.~\ref{fig:high_dim_gauss}  shows that, for all dimensions, both mGradNets closely approximate the optimal transport map and the mGradNet-M consistently outperforms the mGradNet-C.

\subsection{Image Morphing as Optimal Transport}
\begin{figure}[htpb]
    \centering

    \begin{subfigure}[b]{0.24\linewidth}
        \includegraphics[width=\linewidth, trim=0 0 0 25, clip]{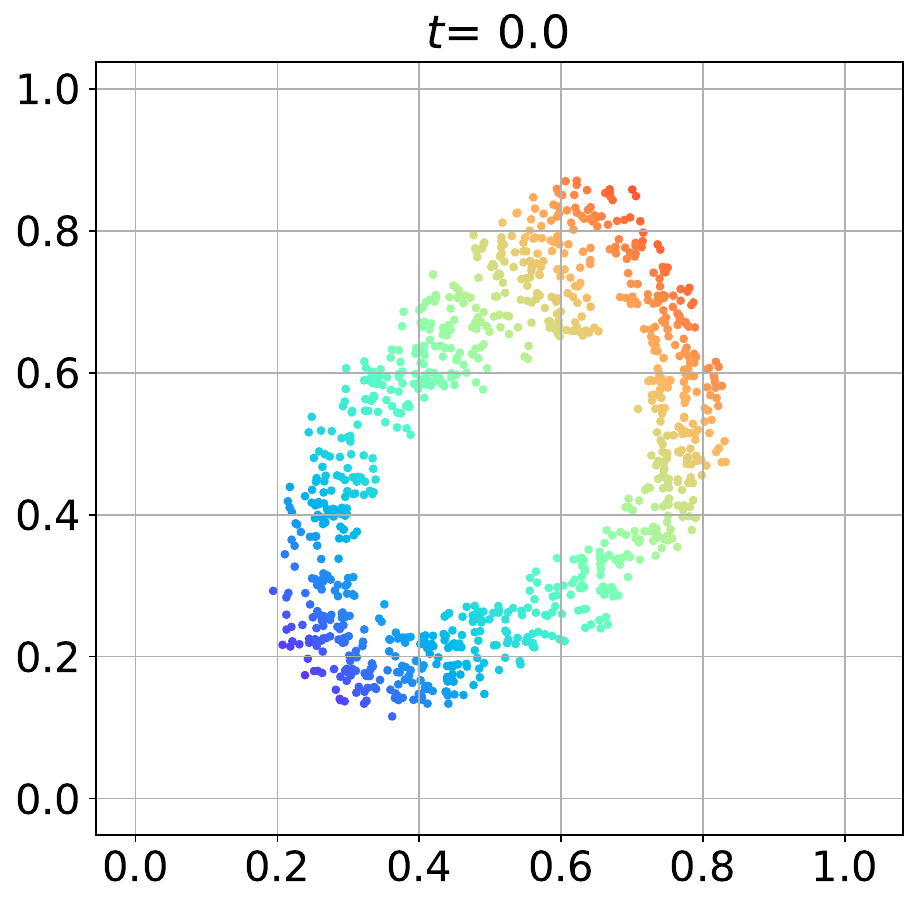}
    \end{subfigure}
    \hfill
    \begin{subfigure}[b]{0.24\linewidth}
        \includegraphics[width=\linewidth, trim=0 0 0 25, clip]{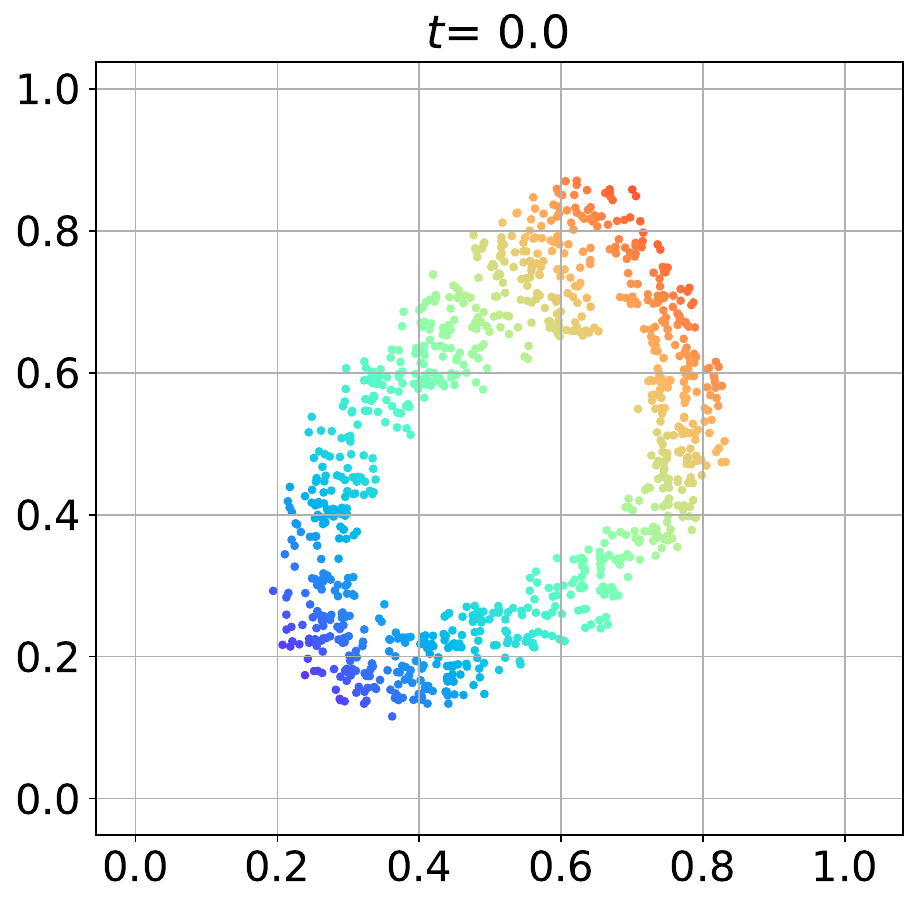}
    \end{subfigure}
    \hfill 
    \begin{subfigure}[b]{0.24\linewidth}
        \includegraphics[width=\linewidth, trim=0 0 0 25, clip]{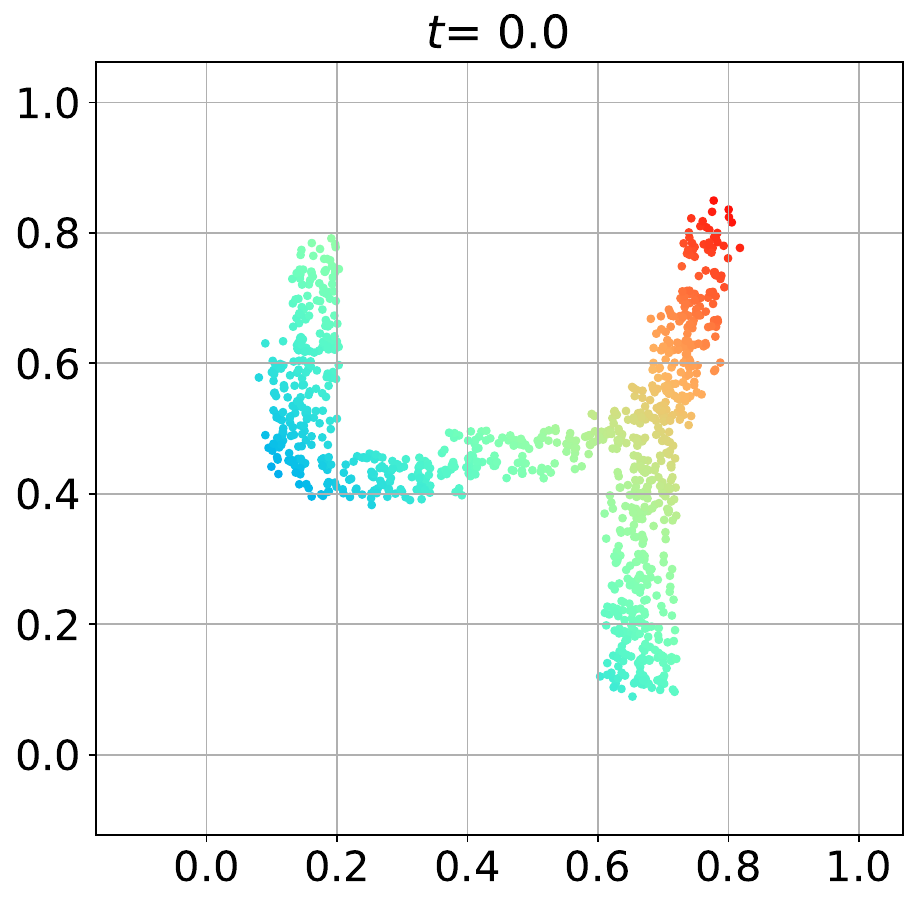}
    \end{subfigure}
    \hfill    
    \begin{subfigure}[b]{0.24\linewidth}
        \includegraphics[width=\linewidth, trim=0 0 0 25, clip]{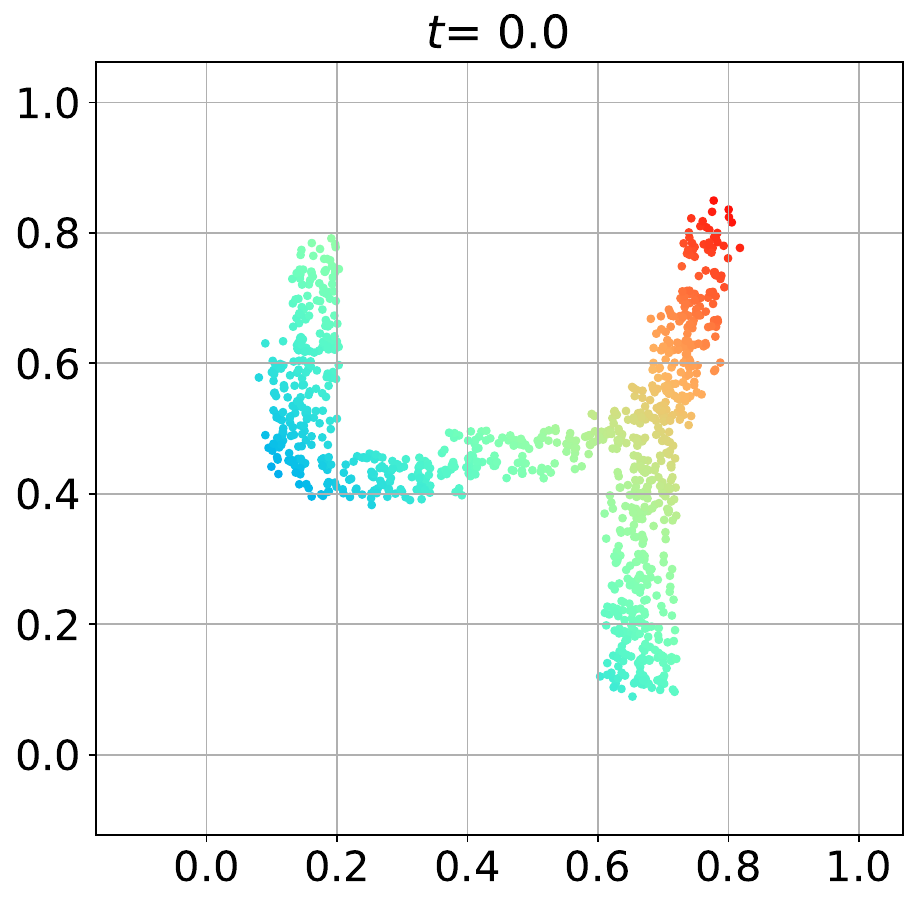}
    \end{subfigure}
    
    \vspace{0.5em}

    \begin{subfigure}[b]{0.24\linewidth}
        \includegraphics[width=\linewidth, trim=0 0 0 25, clip]{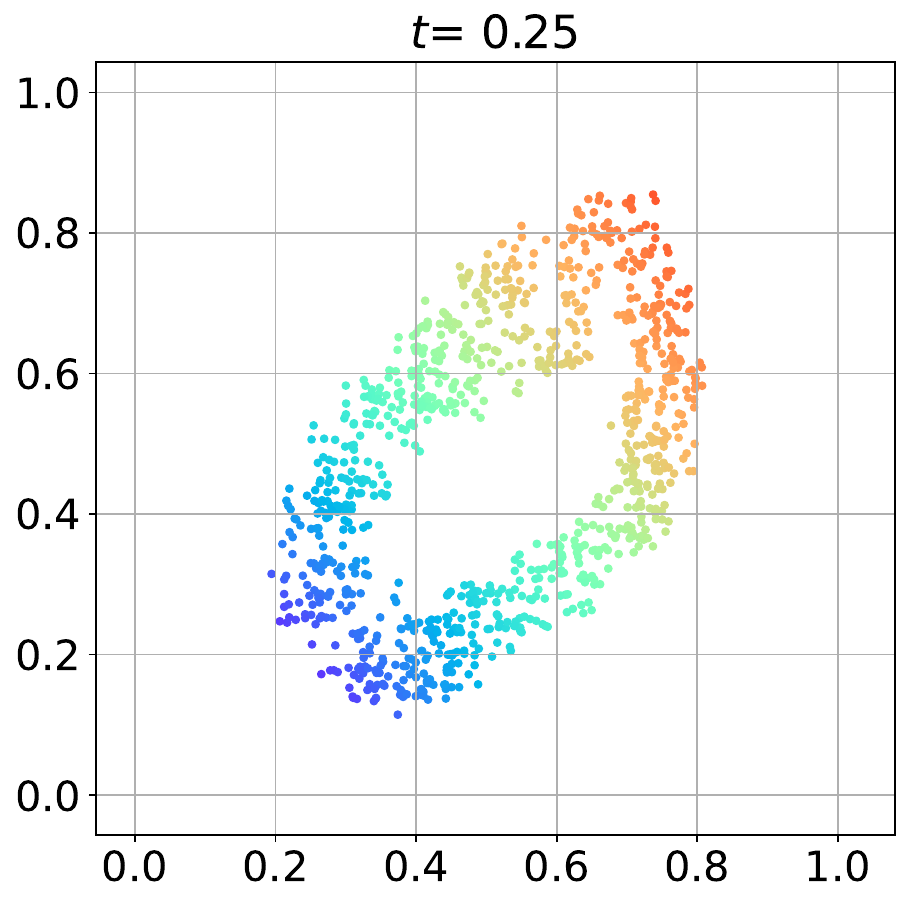}
    \end{subfigure}
    \hfill
    \begin{subfigure}[b]{0.24\linewidth}
        \includegraphics[width=\linewidth, trim=0 0 0 25, clip]{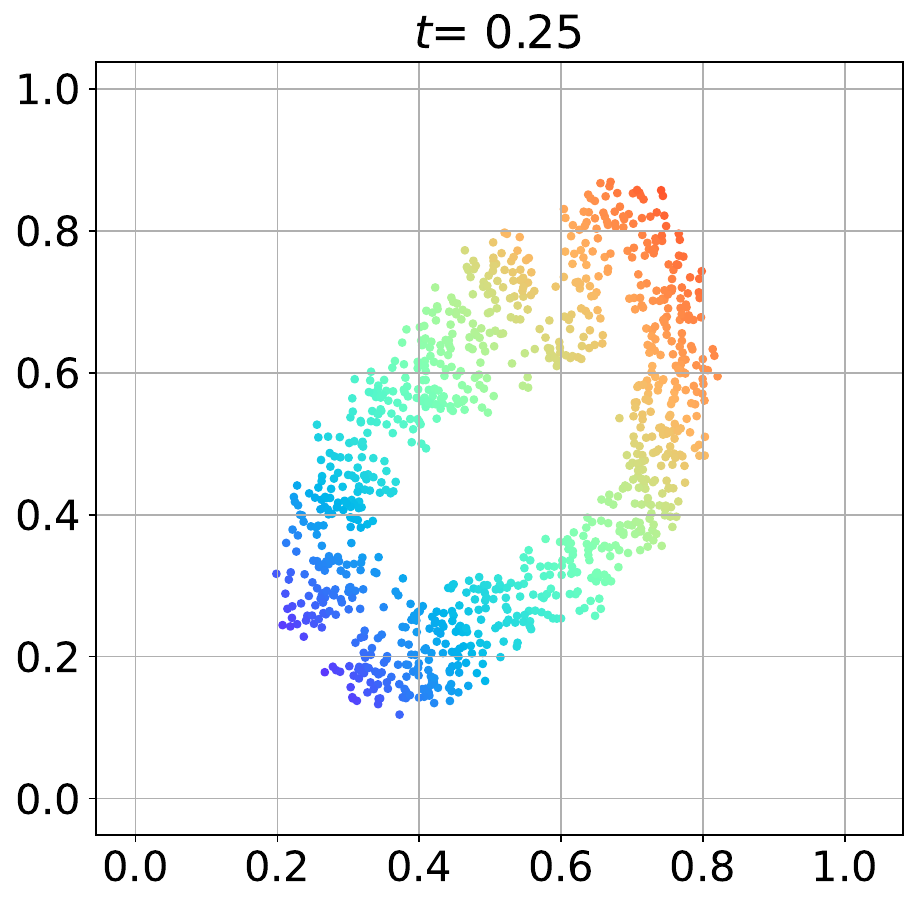}
    \end{subfigure}
    \hfill 
    \begin{subfigure}[b]{0.24\linewidth}
        \includegraphics[width=\linewidth, trim=0 0 0 25, clip]{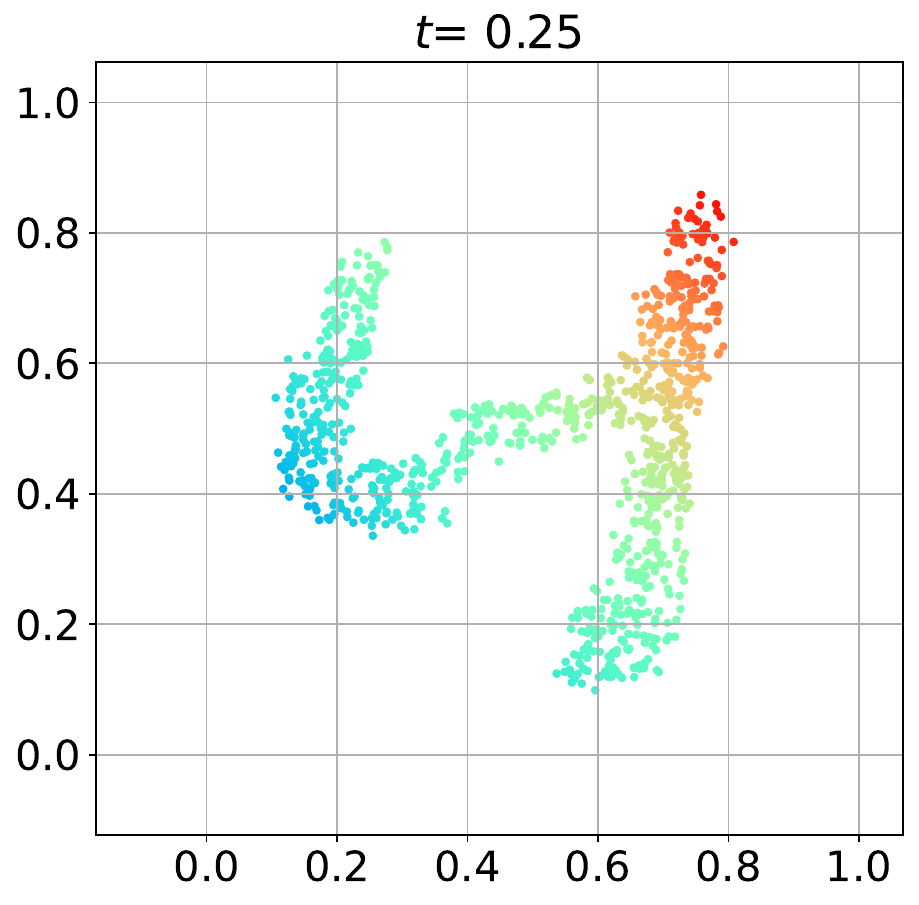}
    \end{subfigure}
    \hfill    
    \begin{subfigure}[b]{0.24\linewidth}
        \includegraphics[width=\linewidth, trim=0 0 0 25, clip]{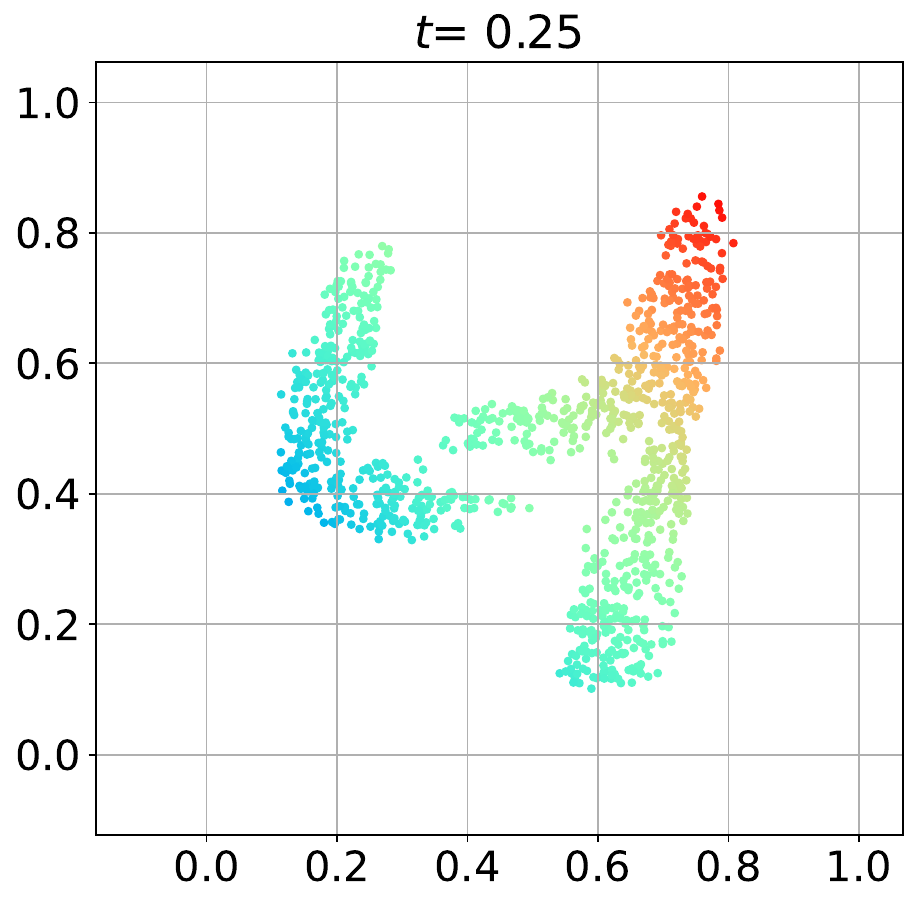}
    \end{subfigure}

    \vspace{0.5em}

    \begin{subfigure}[b]{0.24\linewidth}
        \includegraphics[width=\linewidth, trim=0 0 0 25, clip]{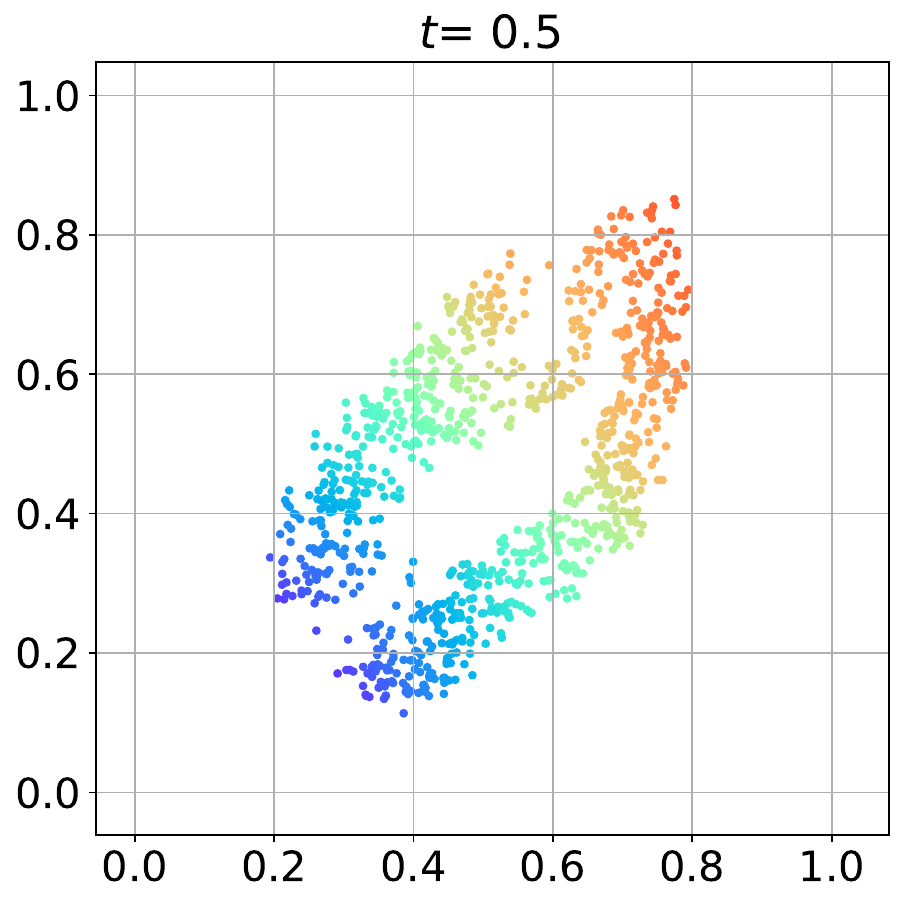}
    \end{subfigure}
    \hfill
    \begin{subfigure}[b]{0.24\linewidth}
        \includegraphics[width=\linewidth, trim=0 0 0 25, clip]{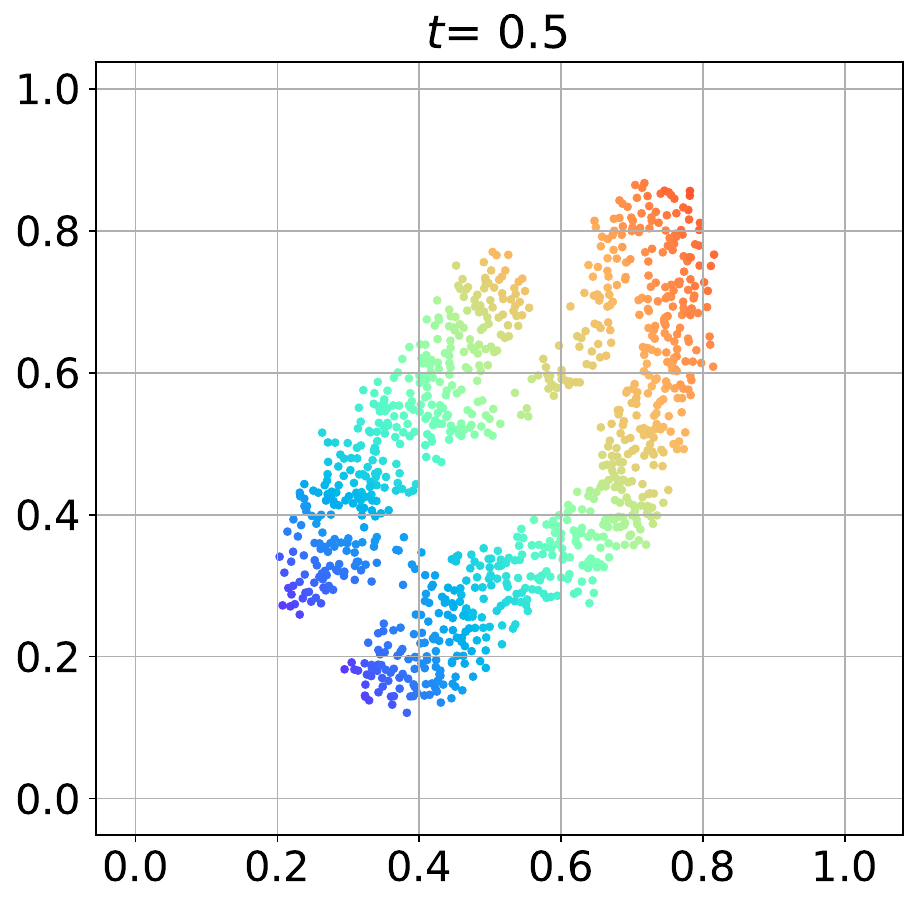}
    \end{subfigure}
    \hfill 
    \begin{subfigure}[b]{0.24\linewidth}
        \includegraphics[width=\linewidth, trim=0 0 0 25, clip]{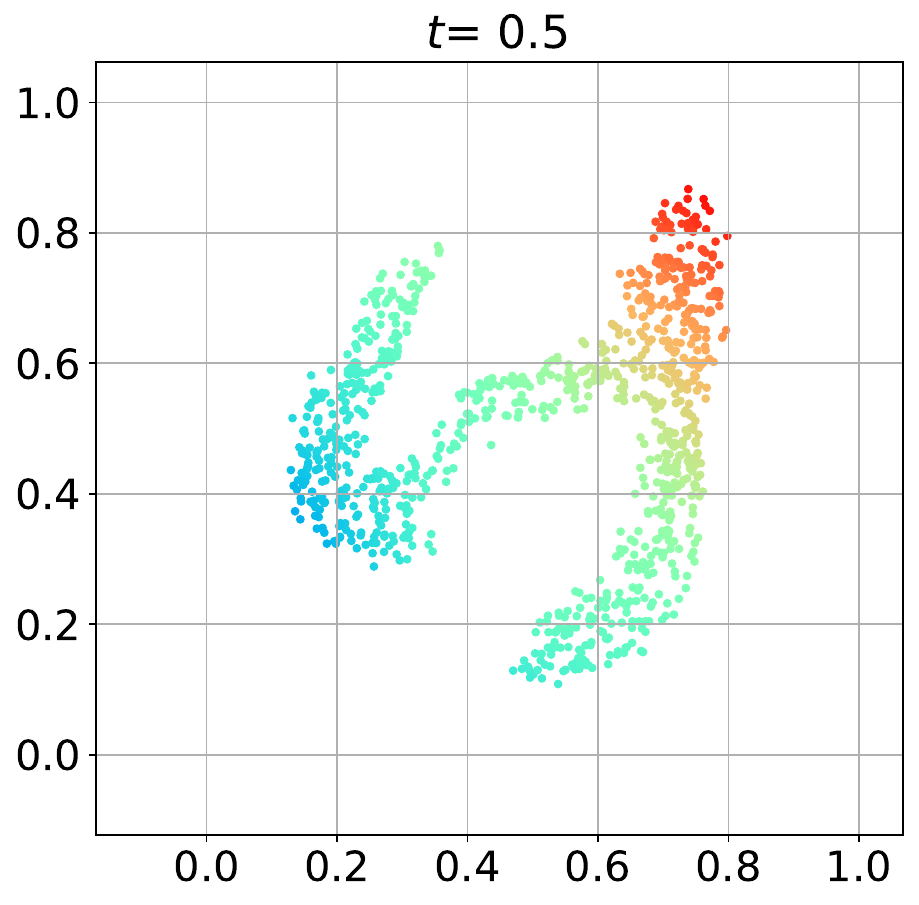}
    \end{subfigure}
    \hfill    
    \begin{subfigure}[b]{0.24\linewidth}
        \includegraphics[width=\linewidth, trim=0 0 0 25, clip]{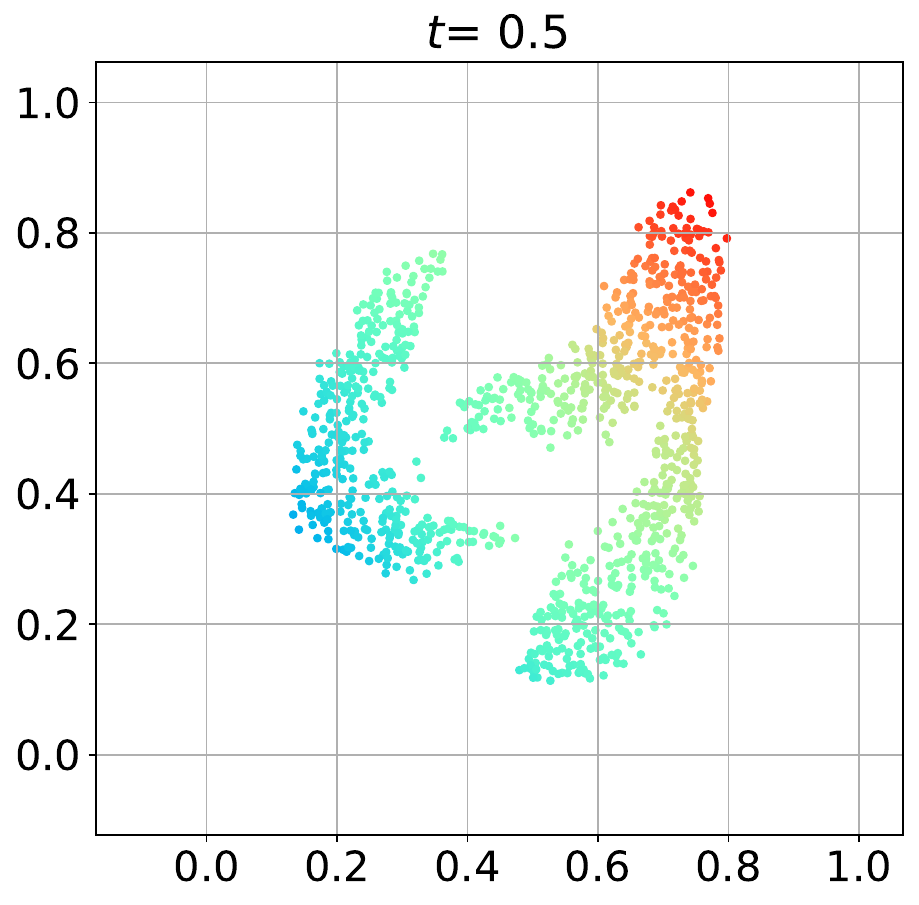}
    \end{subfigure}

    \vspace{0.5em}

    \begin{subfigure}[b]{0.24\linewidth}
        \includegraphics[width=\linewidth, trim=0 0 0 25, clip]{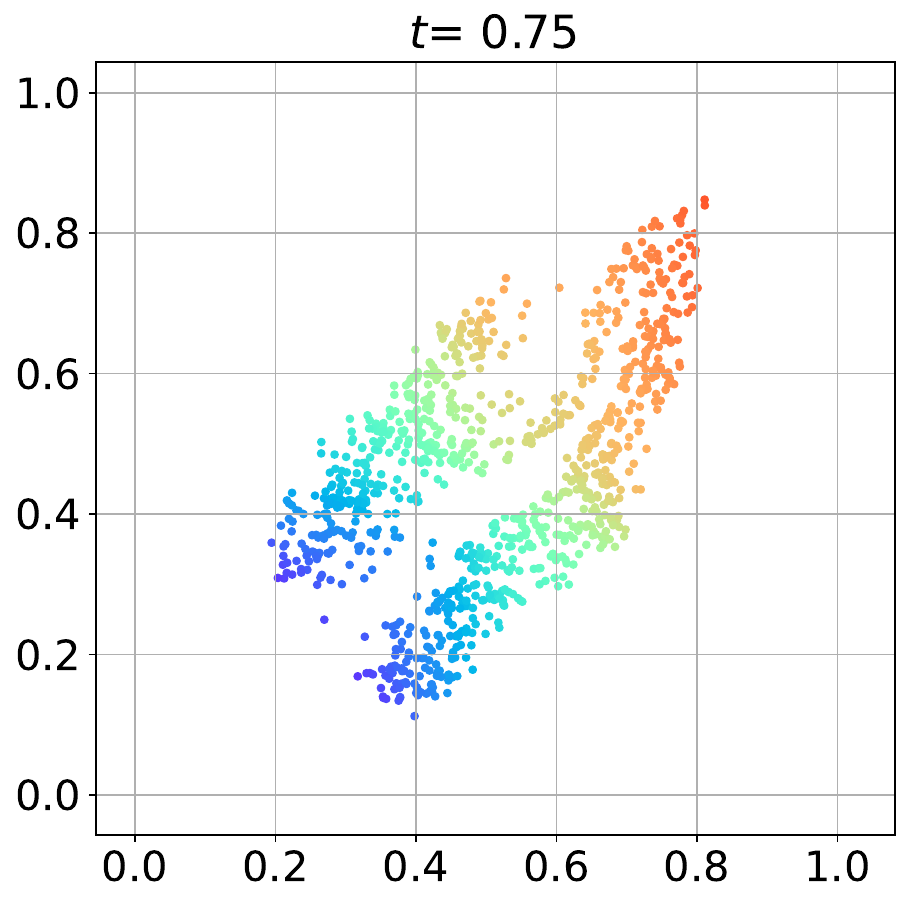}
    \end{subfigure}
    \hfill
    \begin{subfigure}[b]{0.24\linewidth}
        \includegraphics[width=\linewidth, trim=0 0 0 25, clip]{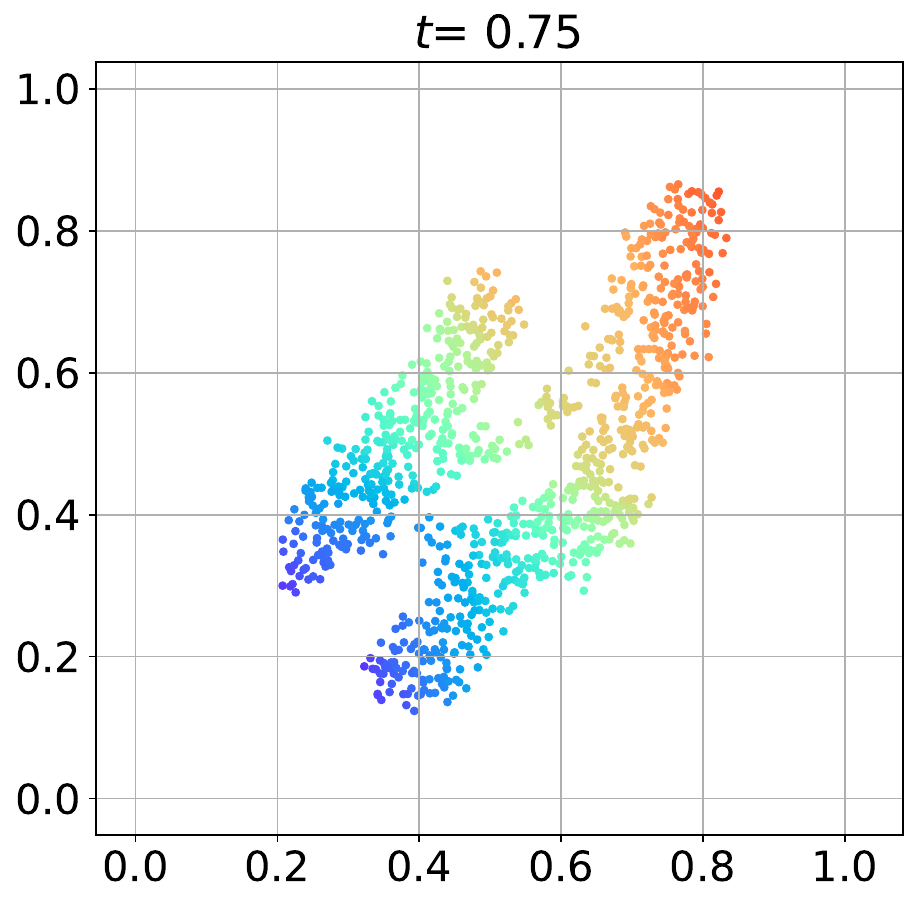}
    \end{subfigure}
    \hfill 
    \begin{subfigure}[b]{0.24\linewidth}
        \includegraphics[width=\linewidth, trim=0 0 0 25, clip]{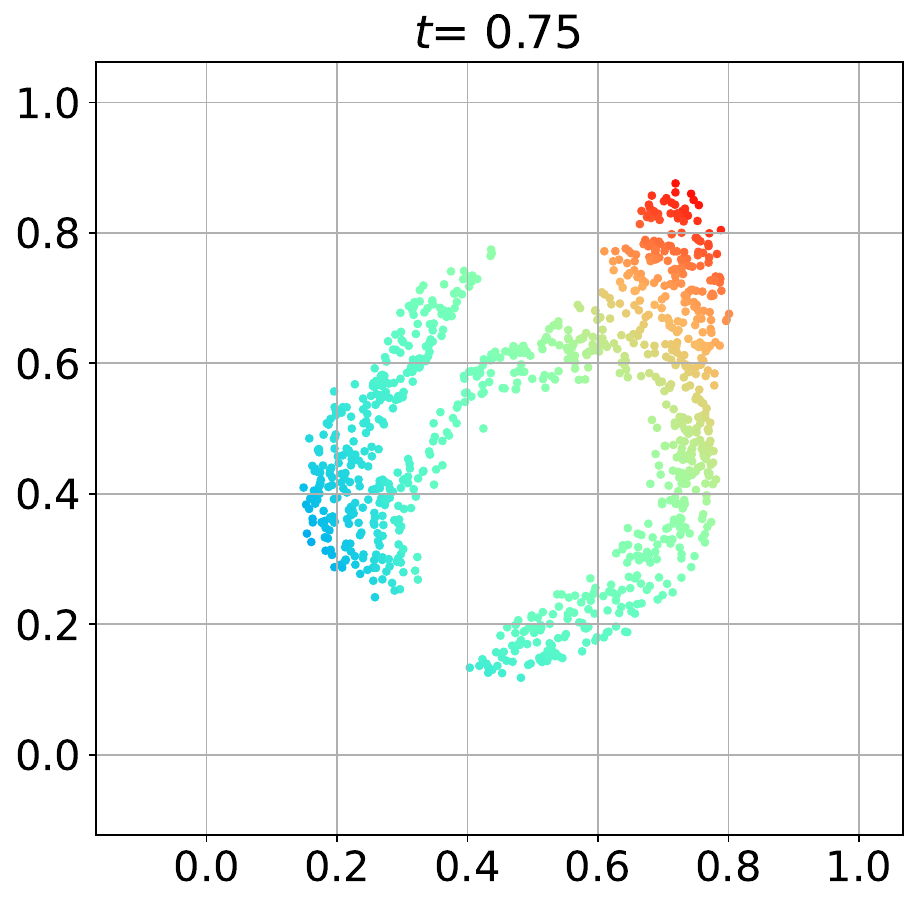}
    \end{subfigure}
    \hfill    
    \begin{subfigure}[b]{0.24\linewidth}
        \includegraphics[width=\linewidth, trim=0 0 0 25, clip]{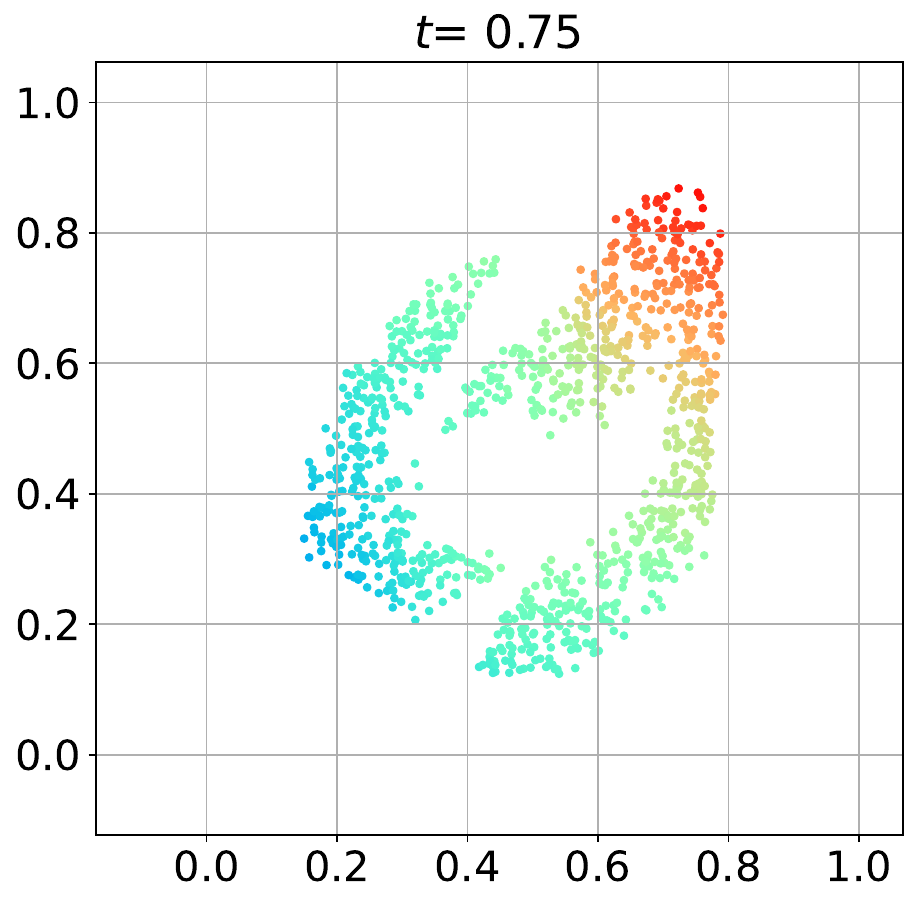}
    \end{subfigure}

    \vspace{0.5em}

    \begin{subfigure}[b]{0.24\linewidth}
        \includegraphics[width=\linewidth, trim=0 0 0 25, clip]{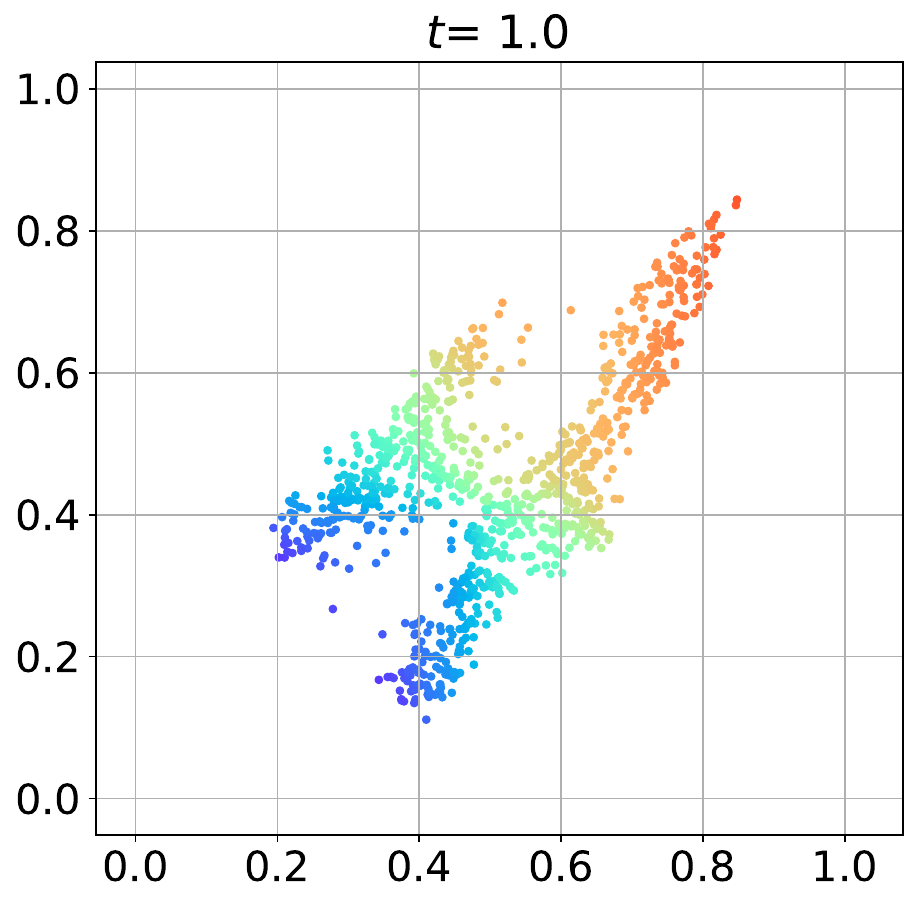}
    \end{subfigure}
    \hfill
    \begin{subfigure}[b]{0.24\linewidth}
        \includegraphics[width=\linewidth, trim=0 0 0 25, clip]{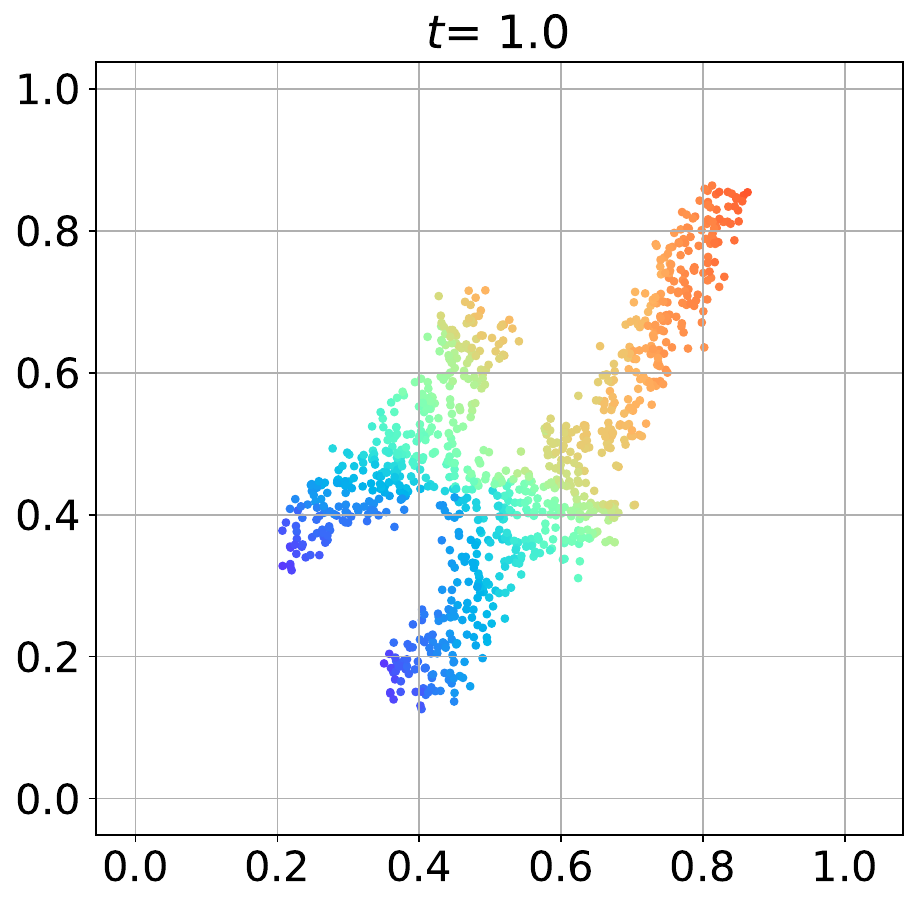}
    \end{subfigure}
    \hfill 
    \begin{subfigure}[b]{0.24\linewidth}
        \includegraphics[width=\linewidth, trim=0 0 0 25, clip]{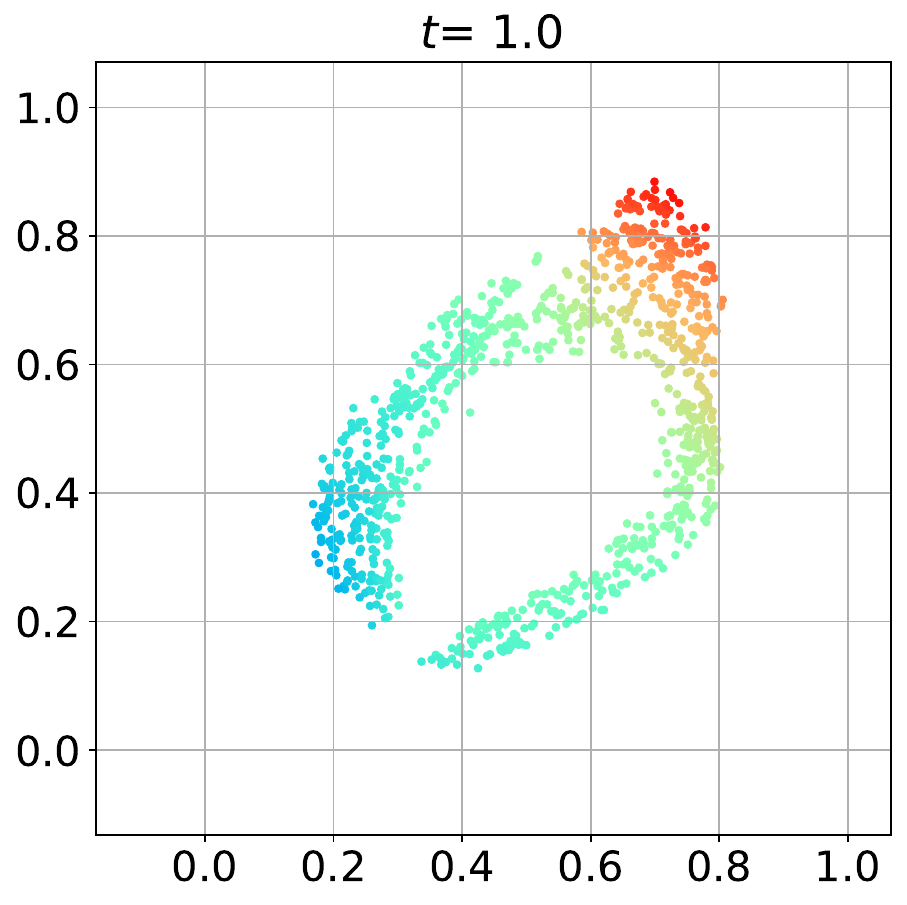}
    \end{subfigure}
    \hfill    
    \begin{subfigure}[b]{0.24\linewidth}
        \includegraphics[width=\linewidth, trim=0 0 0 25, clip]{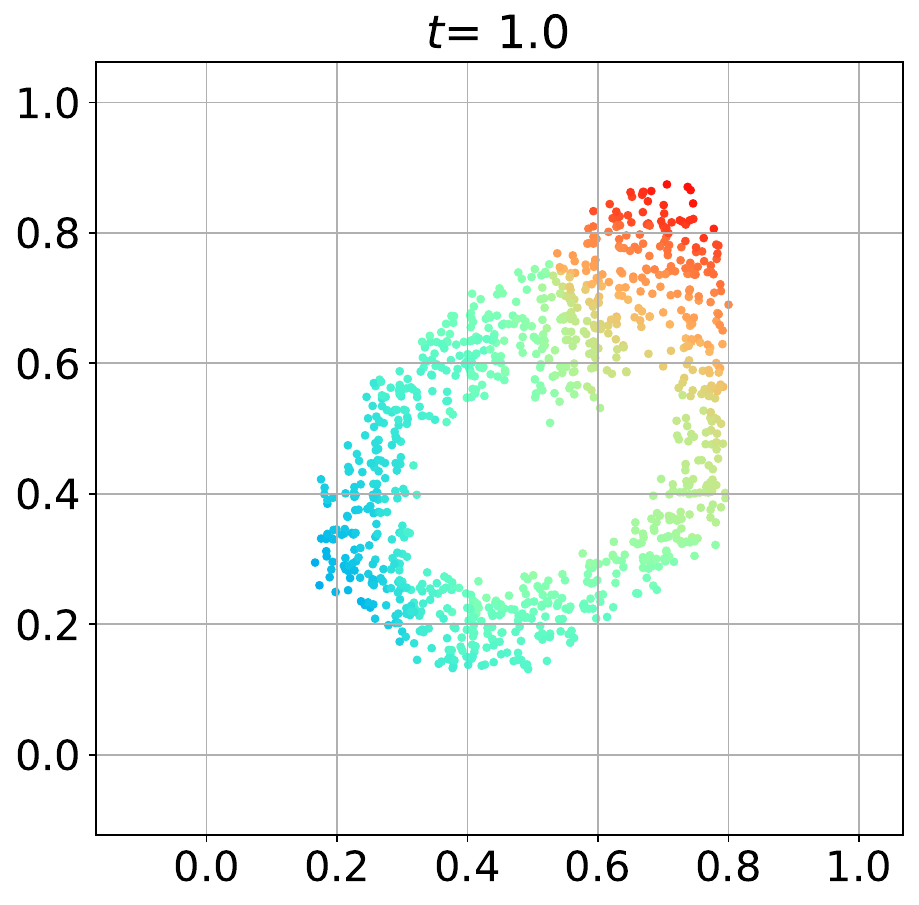}
    \end{subfigure}

    \caption{Left two columns: OT and barycentric interpolation from 0 to 4. Right two columns: from 4 to 0. Rows correspond to different interpolation times \(t \in \{0.0, 0.25, 0.5, 0.75, 1.0\}\).}
    \label{fig:interpolation_04_40}
\end{figure}

We consider the problem of optimally transporting a distribution over pixel locations from their known initial locations to desired locations, thereby effectively morphing the image. We model image pixel intensities from the MNIST dataset as samples from a continuous density over the unit square $[0,1]^2$. Each pixel is treated as a point mass located on a uniform grid and we fit a smooth density using kernel density estimation (KDE). Given an $n \times n$ image with intensity values $I(i,j)$, we model the density over pixel locations $\x \in [0,1]^2$ as the following mixture of Gaussians:
\begin{align}
    p(\x) = \sum_{i=0}^{n-1} \sum_{j=0}^{n-1} I(i, j) \cdot \mathcal{N}\left(\x ; \left(\frac{i}{n-1}, \frac{j}{n-1}\right), \sigma^2 \mathbf{I} \right)
\end{align}
with $\sigma^2 = 10^{-4}$ and $n = 28$ for MNIST. We train an mGradNet-M to learn the optimal transport map between these densities using Algorithm~\ref{alg:main_alg}. During training, we exponentially decay the learning rate from 1e-2 to 1e-4 over 10000 iterations and also decay $\sigma^2$ in the same manner to facilitate more stable training. In Fig.~\ref{fig:interpolation_04_40}, we illustrate the learned OT displacement map 
\begin{align}
    \x^{(t)} = (1-t)\x^{(0)} + tT_{\boldsymbol{\theta}}\left(\x^{(0)}\right)
\end{align}
for morphing between a 0 and a 4.
The first column corresponds to the transport map learned by the mGradNet-M. The second column corresponds to the transport map from the barycentric projection. The barycentric projection converts a discrete optimal transport plan, from solving the Monge-Kantorovich relaxation, to an optimal transport map. Given an optimal transport plan $\gamma \in \R^{N_x \times N_y}$, which describes how much mass to move from source points $\{\x_i\}_{i=1}^{N_x}$ to target points $\{\y_j\}_{j=1}^{N_y}$, the barycentric projection described in \cite{peyre2019computational} is defined as: 
\begin{align}
    T(\x_i) = \frac{\sum_{j=1}^{N_y} \gamma_{ij}\y_j}{\sum_{j=1}^{N_y}\gamma_{ij}}
\end{align}
Specifically, the barycentric projection approximates a Monge (not mass splitting) transport map using a potentially mass-splitting Kantorovich plan. When the solution to the Kantorovich problem $\gamma$ does not split mass, and a Monge map exists, then the barycentric projection recovers the optimal Monge map in the limit as the number of source and target points increases \cite{peyre2019computational, villani2009optimal}. However, computing the optimal transport plan scales quadratically with the number of points, motivating the learning of a parametric OT map, which avoids the cost of solving and storing a full discrete transport plan.
\begin{table}[htbp]
    \centering
    \caption{MSE between learned OT map and barycentric projection for digits 0--4 from MNIST.}
    \label{tab:bary_results}
    \renewcommand{\arraystretch}{1.2}
    \begin{tabular}{|c|c|c|c|c|c|c|}
    \hline
    & & \multicolumn{5}{c|}{\textbf{Target}} \\ \cline{3-7}
    & & \textbf{0} & \textbf{1} & \textbf{2} & \textbf{3} & \textbf{4} \\
    \hline
    \multirow{5}{*}{\rotatebox{90}{\textbf{Source}}}
    & \textbf{0} & \cellcolor{green!30}0.00138 & \cellcolor{green!10}0.00031 & \cellcolor{green!90}0.00566 & \cellcolor{green!20}0.00036 & \cellcolor{green!40}0.00145 \\
    & \textbf{1} & \cellcolor{green!35}0.00146 & \cellcolor{green!5}0.00024  & \cellcolor{green!70}0.00438 & \cellcolor{green!55}0.00178 & \cellcolor{green!30}0.00104 \\
    & \textbf{2} & \cellcolor{green!25}0.00123 & \cellcolor{green!15}0.00038 & \cellcolor{green!30}0.00146 & \cellcolor{green!25}0.00055 & \cellcolor{green!25}0.00055 \\
    & \textbf{3} & \cellcolor{green!20}0.00117 & \cellcolor{green!7}0.00026  & \cellcolor{green!45}0.00253 & \cellcolor{green!23}0.00051 & \cellcolor{green!30}0.00105 \\
    & \textbf{4} & \cellcolor{green!60}0.00273 & \cellcolor{green!13}0.00034 & \cellcolor{green!28}0.00170 & \cellcolor{green!50}0.00159 & \cellcolor{green!18}0.00082 \\
    \hline
    \end{tabular}
\end{table}

In Table \ref{tab:bary_results} we report the mean squared error (MSE) between the barycentric projection $T(\x_i)$ and the learned mapping $T_{\boldsymbol{\theta}}(\x_i)$ over 1000 samples $\x_i$ for various source and target image pairs.
Each entry in the table reports the difference between the transport map learned by an mGradNet-M and the barycentric projection, when transporting mass from the source density (row) to the target density (column). Lower MSE values in the table imply that the transport cost of the learned OT map is closer to that of the optimal transport map. Overall, we observe both numerically and qualitatively in Fig.~\ref{fig:interpolation_04_40}, that our proposed method closely matches the performance of the barycentric projection.

\section{Conclusion}
We propose a method for learning optimal transport maps using mGradNets -- neural networks guaranteed to correspond to gradients of convex functions. Inspired by Brenier’s theorem, our approach finds solutions to the Monge problem with quadratic Euclidean cost by directly searching over the space of gradients of convex potentials by optimizing a Monge-Amp\`ere loss. We validate our method through a series of experiments demonstrating its effectiveness. Our framework opens avenues for further research in the design and training of theoretically-motivated neural networks for solving optimal transport problems.

\newpage 
\bibliographystyle{IEEEbib}
\bibliography{refs}

\end{document}